\documentclass[%
 aip,%
 amsmath,amssymb,%
 reprint,%
]{revtex4-2}

\usepackage[utf8]{inputenc}
\usepackage[T1]{fontenc}
\usepackage{mathptmx}
\usepackage{graphicx}
\usepackage{booktabs}
\usepackage{multirow}
\usepackage{threeparttable}
\usepackage{url}

\begin{document}

\title{HGPTrans: Hierarchical Graph-Pooling Transolver for Automotive
      Aerodynamic Drag Coefficient Prediction}

\author{Bo Liu}
\email{boliu128@mail.ustc.edu.cn}
\affiliation{Institute of Auto Engineering, BYD Auto Industry Co., Ltd., Shenzhen, China}

\author{Fengli Zhang}
\affiliation{Institute of Auto Engineering, BYD Auto Industry Co., Ltd., Shenzhen, China}
\affiliation{Center for Turbulence Control, Harbin Institute of Technology, Shenzhen, China}

\author{Qiuli Luo}
\affiliation{Institute of Auto Engineering, BYD Auto Industry Co., Ltd., Shenzhen, China}

\author{Lianrui Nie}
\affiliation{Institute of Auto Engineering, BYD Auto Industry Co., Ltd., Shenzhen, China}

\author{Wenjiang Wang}
\affiliation{Institute of Auto Engineering, BYD Auto Industry Co., Ltd., Shenzhen, China}

\date{\today}

\begin{abstract}Accurate and rapid prediction of the aerodynamic drag coefficient (\(C_D\)) is essential for vehicle design, particularly during early-stage design, where many candidate geometries must be evaluated. Although computational fluid dynamics (CFD) provides reliable aerodynamic estimates, its high computational cost limits large-scale design exploration. This paper proposes the hierarchical graph-pooling Transolver (HGPTrans), which combines hierarchical graph pooling with Transolver-based attention to directly predict \(C_D\) from vehicle surface meshes. Motivated by the fact that vehicle aerodynamics depends on both local geometric features and long-range interactions among spatially distant surface regions, HGPTrans integrates three complementary components. Graph isomorphism convolutions encode discriminative local geometry, Transolver-style slice attention captures global interactions with linear computational complexity, and information-redundancy-aware hierarchical pooling progressively removes redundant nodes while preserving informative geometric structures. The model is trained and evaluated on the large-scale DrivAerNet and DrivAerNet++ datasets, where it achieves the lowest mean absolute error and mean squared error among the evaluated baselines. Its generalization capability is further assessed through transfer learning on a real-vehicle dataset containing both sedans and sport utility vehicles (SUVs), achieving relative \(L_1\) errors of 1.56\% (sedans) and 2.12\% (SUVs) with an inference time of approximately \(0.293\) s per vehicle. This corresponds to an acceleration of several orders of magnitude relative to high-fidelity CFD while keeping the predicted drag coefficients within a few percent of the CFD reference. Ablation studies confirm each component's contribution and reveal the effects of depth and pooling ratio.
\end{abstract}

\keywords{Drag coefficient prediction, Graph neural network, Hierarchical graph pooling, Transfer learning}

\maketitle

\begin{center}
  \begin{minipage}{0.98\linewidth}
    \centering
    {\small \textbf{Abbreviations used in this paper.}}\par\vspace{2pt}
    {\small
    \begin{tabular}{@{}ll@{}}
      \toprule
      Abbreviation & Full form \\
      \midrule
      $C_D$ & Drag coefficient \\
      CFD & Computational fluid dynamics \\
      GIN & Graph isomorphism network \\
      GCN & Graph convolutional network \\
      QEM & Quadric error metric \\
      MAE & Mean absolute error \\
      MSE & Mean squared error \\
      MLP & Multilayer perceptron \\
      \bottomrule
    \end{tabular}}
  \end{minipage}
\end{center}

\section{Introduction}\label{sec:intro}

Aerodynamic performance is a primary factor in the energy efficiency and
dynamic stability of road vehicles. The aerodynamic drag coefficient $C_D$
directly determines the aerodynamic drag force acting on the vehicle body, and
even modest reductions in $C_D$ can improve the driving range of an electric
vehicle and the fuel economy of a conventional vehicle \cite{ref_sudin}.
Consequently, the accurate evaluation of $C_D$
constitutes a core task in the vehicle aerodynamic design loop, in which
stylists and aerodynamicists iterate over a large number of shape variants
within a tightly constrained development schedule \cite{ref_hucho,ref_jacob}.

The conventional approach to evaluating $C_D$ relies on wind-tunnel
experiments and computational fluid dynamics (CFD) simulations. Wind-tunnel
tests require physical prototypes and are therefore expensive and slow. CFD
simulations, which resolve the turbulent flow field around the vehicle body,
provide a reliable estimate of the drag coefficient together with a detailed
surface pressure and wall shear stress distribution \cite{ref_argyropoulos}.
However, a single high-fidelity CFD evaluation demands geometry clean-up, mesh
generation, numerical solution, and post-processing, and typically consumes
hours to days of computational time for a production-scale surface mesh
comprising on the order of $10^6$--$10^7$ faces, which corresponds to a
volume mesh of tens of millions to hundreds of millions of cells
\cite{ref_hucho,ref_faster}. When dozens to
hundreds of design candidates must be screened during the styling phase, the
accumulated cost of CFD becomes a severe bottleneck that limits the size of the
design space that can be explored \cite{ref_jacob,ref_drivaerml}.

Driven by the increasing availability of large-scale automotive CFD datasets
\cite{ref_drivaernet,ref_drivaernetpp,ref_drivaerml,ref_drivaerstar}, deep
learning has emerged as a promising alternative for the fast prediction of
aerodynamic coefficients. Given a geometric representation of the vehicle
body, a neural surrogate model is trained to regress $C_D$ directly, thereby
shifting the computational burden from expensive online simulation to a
one-time offline training phase \cite{ref_jacob,ref_gnot,ref_fno}. Existing
surrogates differ mainly in the way the vehicle geometry is represented and
processed. Voxel-based convolutional networks discretize the geometry onto a
regular grid and apply volumetric convolution, but suffer from an
$O(N^3)$ computational complexity and a loss of fine surface detail
\cite{ref_jacob,ref_voxnet}. Mesh-based convolutional networks instead
operate directly on the surface mesh elements and thereby preserve the
surface connectivity \cite{ref_meshcnn}, but their pooling is defined on
the mesh elements rather than on the graph vertices. Depth-and-normal
rendering approaches project the three-dimensional geometry onto a
two-dimensional image and predict $C_D$ from the rendered views
\cite{ref_song}. Point-cloud networks such as PointNet \cite{ref_pointnet}
operate directly on unordered point sets and disregard the explicit
connectivity of the surface mesh, although they remain widely used for
automotive drag prediction \cite{ref_drivaernet}. In contrast, graph
neural networks (GNNs) represent the vehicle surface as a graph whose nodes
carry the vertex coordinates and surface normals, and exploit the mesh
adjacency to aggregate local geometric information
\cite{ref_gin,ref_meshgraphnets,ref_gns}. This representation retains the
explicit topology of the surface and is well suited to the irregular,
non-uniform meshes produced by automotive CFD \cite{ref_drivaernet,ref_aerogto}.

Despite these advances, three challenges remain. First, the aerodynamic
behavior of a vehicle is highly sensitive to local topological details: small
geometric variations at front-flow deflectors, side mirrors, underbody
channels, and wheel arches can markedly change the drag coefficient
\cite{ref_drivaernet,ref_sudin}. A surrogate model must therefore encode such
local surface features with high fidelity. Second, the drag coefficient is a
global integral quantity that results from the coupled interaction of
pressure waves, vortices, and wake structures distributed over the entire
body; purely local message passing cannot capture these long-range
dependencies without a large number of propagation steps
\cite{ref_meshgraphnets,ref_gns,ref_aerogto}.
Third, production-scale vehicle meshes contain millions of vertices, and a
direct global attention over all vertices incurs an $O(N^2)$ cost that is
computationally prohibitive \cite{ref_transolver,ref_transolver3}. Accordingly,
an effective drag prediction framework must jointly preserve local geometry,
model global aerodynamic coupling, and remain affordable on large meshes.

Several recent works have made progress toward these objectives. AeroGTO
\cite{ref_aerogto} combines graph message passing with an efficient global
attention operator to predict the surface pressure field and the drag of
large-scale vehicle geometries, yet it processes the geometry at a single
resolution and does not explicitly coarsen the graph. DoMINO \cite{ref_domino}
builds a multi-scale iterative neural operator on point clouds and demonstrates
strong performance on the DrivAerML dataset, but it relies on point-cloud
representations that discard the explicit surface connectivity. Transolver
\cite{ref_transolver}, Transolver++ \cite{ref_transolverplusplus}, and
Transolver-3 \cite{ref_transolver3} advance physics-aware slice attention to
industrial-scale meshes of up to hundreds of millions of cells, but the
primary focus of these methods is the reconstruction of high-resolution flow fields rather than
the graph-level regression of a single scalar coefficient under a strict
computational budget. Similarly, projection- and triplane-based methods such as
FIGConv \cite{ref_figconv} and TripNet \cite{ref_tripnet} achieve compelling
accuracy on automotive CFD benchmarks, but the implicit or projected
representations adopted by these methods do not directly leverage the
hierarchical redundancy that is inherent in a surface mesh. These observations motivate the development of a
hierarchical graph-based framework that explicitly coarsens the mesh while
preserving the aerodynamic-critical geometry, and that is evaluated not only on
public benchmarks but also on production vehicle data through a dedicated
cross-domain transfer procedure.

To address these requirements, this paper proposes the hierarchical
graph-pooling Transolver (HGPTrans), which combines hierarchical graph pooling
with a Transolver-style slice attention mechanism, for the end-to-end
prediction of the drag coefficient from the
vehicle surface mesh. The main contributions are as follows:

\begin{enumerate}
  \item \textbf{A local--global--hierarchical architecture.} HGPTrans
        decomposes the drag prediction task into three complementary stages.
        A graph isomorphism convolution (GIN) aggregates local geometric
        features from the mesh adjacency, a Transolver slice attention models
        the global aerodynamic interaction between distant vehicle regions in
        a linear computational budget, and an information-redundancy-aware
        hierarchical pooling progressively coarsens the graph while retaining
        the geometrically informative vertices.
  \item \textbf{An information-score-based pooling criterion.} A pooling ratio
        is used to retain vertices whose features have relatively large
        neighborhood-reconstruction errors, thereby reducing redundant mesh
        subdivisions while retaining the high-scoring geometric features.
  \item \textbf{Cross-domain transfer learning.} The model is pre-trained on
        the large-scale DrivAerNet++ dataset and subsequently fine-tuned on a
        real-vehicle dataset, assessing how the representation learned from
        synthetic parametric geometries can be adapted to production vehicle
        bodies.
  \item \textbf{Comprehensive evaluation.} Extensive experiments are conducted
        on DrivAerNet, DrivAerNet++, and a real-vehicle dataset that covers
        both sedans and sport utility vehicles (SUVs), together with architecture and hyperparameter
        ablation studies.
\end{enumerate}

The remainder of this paper is organized as follows. Section~\ref{sec:method}
presents the overall architecture of HGPTrans and details its three core
components. Section~\ref{sec:dataset} introduces the benchmark and real-vehicle
datasets and the training strategy. Section~\ref{sec:experiments} reports the
experimental results, including the comparison with mainstream baselines,
transfer learning on real vehicles, and ablation studies. Finally,
Section~\ref{sec:conclusions} concludes the paper.

\section{Method}\label{sec:method}

\subsection{Mesh preprocessing and overall architecture}\label{sec:architecture}

\begin{figure*}[tb]
  \centering
  \includegraphics[width=\linewidth]{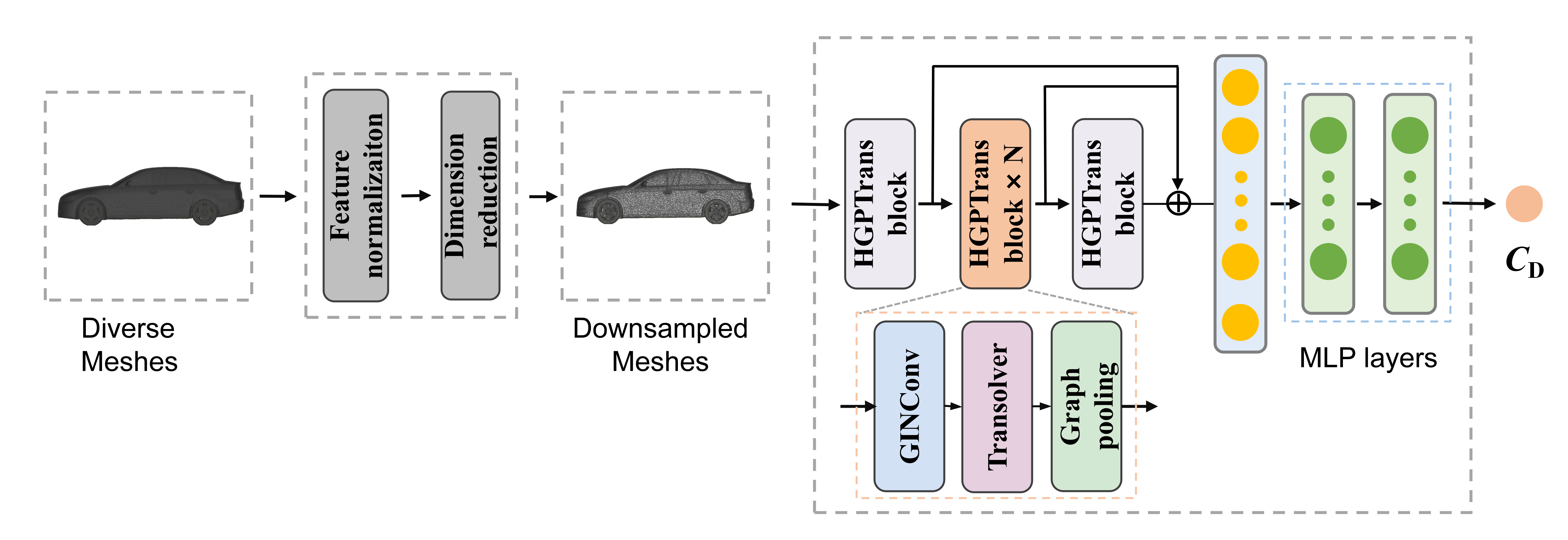}
  \caption{The overall architecture of HGPTrans. The input
           vehicle meshes are first normalized, and high-resolution meshes are
           simplified using QEM to obtain downsampled meshes. The processed mesh
           is converted into a graph and passed through the stacked HGPTrans
           blocks, graph-level pooling, and fully connected layers to predict
           the drag coefficient $C_D$.}
  \label{fig:arch}
\end{figure*}

The complete processing pipeline begins with geometric normalization and mesh
resolution reduction, as illustrated in Fig.~\ref{fig:arch}. The vertex
coordinates are first normalized to remove differences in coordinate scale and
to provide a consistent input representation. Meshes that exceed the target
resolution, principally the dense real-vehicle meshes, are then simplified
using the quadric error metric (QEM) algorithm \cite{ref_qem}. QEM iteratively
contracts the vertex pairs that introduce the smallest geometric error,
producing a downsampled mesh with substantially fewer vertices while retaining
the principal vehicle shape and local surface characteristics. Benchmark
meshes that are already at the target resolution do not require this additional
simplification step.

After preprocessing, a vehicle surface is represented by a triangular mesh
$\mathcal{M}=(\mathbf{V},\mathbf{E})$, where
$\mathbf{V}\in\mathbb{R}^{N\times 3}$ denotes the vertex coordinates and
$\mathbf{E}$ encodes the edge adjacency of the surface. The task is to learn a
mapping $f:\mathcal{M}\mapsto C_D\in\mathbb{R}$ that regresses the drag
coefficient directly from the processed mesh geometry. The mesh is converted
into a graph in which each vertex becomes a node and each mesh edge becomes a
graph edge. The node feature matrix
$\mathbf{X}\in\mathbb{R}^{N\times 6}$ stacks the normalized three-dimensional
coordinates and the three components of the surface normal, which provide both
the spatial position and the local orientation of each vertex.

Following the normalization and QEM-based downsampling stages, the processed
mesh enters the HGPTrans network shown in Fig.~\ref{fig:arch}. The network
consists of a shape-compression subnetwork and a drag-regression head. The
shape-compression subnetwork stacks $L$ consecutive HGPTrans blocks. Each block
comprises three layers: a graph isomorphism convolution that extracts local
 geometric features, a Transolver-style slice attention layer that captures
 the global aerodynamic interaction, and a hierarchical graph-pooling layer that
 progressively coarsens the graph. At each block, global mean and max pooling
 are applied to the retained node features; the resulting block-level
 embeddings are summed across the $L$ blocks and passed to a fully connected
 regression head that outputs the scalar drag coefficient $\hat{C}_D$.

The hierarchical structure acts as a natural multi-scale representation of the
vehicle. In the early blocks, the graph retains a large number of vertices and
therefore captures the fine local geometry; as the graph is progressively
coarsened, the retained nodes represent a sparser set of informative surface
locations while the global attention maintains the coupling among the regions. This design reduces
the computational cost of the subsequent attention layers while preserving the
aerodynamic-relevant information, and it is directly motivated by the
observation that the drag coefficient depends on both the local surface detail
and the global body shape.

\subsection{Graph convolution with a graph isomorphism network}\label{sec:gin}

The local geometric features of the vehicle surface are extracted by a graph
convolution based on the graph isomorphism network (GIN) proposed in
\cite{ref_gin}. Conventional graph convolutional networks (GCNs) use a
degree-normalized aggregation of self and neighboring features
\cite{ref_gcn}. In contrast, the GIN layer adopts an unnormalized sum
aggregation, which retains the complete information of the neighboring
multiset and is able to distinguish a wider class of local graph structures.
The update rule of a GIN layer reads
\begin{equation}
  \label{eq:gin}
  \mathbf{h}_v^{(\ell+1)} = \mathrm{MLP}\left(
    (1+\epsilon)\cdot \mathbf{h}_v^{(\ell)}
    + \sum_{u\in\mathcal{N}(v)} \mathbf{h}_u^{(\ell)}
  \right),
\end{equation}
where $\mathbf{h}_v^{(\ell)}$ is the feature of vertex $v$ at the input of
layer $\ell$ and $\mathbf{h}_v^{(\ell+1)}$ is the updated feature it produces,
$\mathcal{N}(v)$ denotes the set of its neighbors, and $\epsilon$ is a
learnable scalar that weighs the self-information of the vertex against the
aggregated neighbor information. Because the sum aggregation does not discard
the multiplicity of the neighbor features, the GIN layer can discriminate
local surface patterns, such as corners, edges, and curvature discontinuities,
that are averaged away by mean or max aggregation. This property is
particularly relevant for automotive meshes, in which the aerodynamic-critical
regions are often characterized by sharp local geometric transitions.

The input to the first GIN layer is the normalized node feature matrix
$\mathbf{X}\in\mathbb{R}^{N\times 6}$. The six input channels, comprising the
three vertex-coordinate components and the three surface-normal components,
are standardized independently for each mesh to zero mean and unit variance.
A linear projection maps these six
scalar features into a hidden representation of dimension $d$, after which the
GIN update in Eq.~\eqref{eq:gin} is applied. The learned features of the graph
convolution thereby fuse the coordinate and normal information with the
explicit topology of the surface mesh, providing a local geometric encoding
for the subsequent global attention stage.

\subsection{Global interaction with Transolver-style slice
attention}\label{sec:transolver}

The receptive field of a graph convolution is limited by the number of stacked
layers, so that a purely convolutional network must be very deep to propagate
information across the entire vehicle body. The drag coefficient, however, is
determined by the long-range coupling of flow regions distributed over the
whole vehicle, and a global interaction mechanism is required. A standard
self-attention over all $N$ vertices would incur an $O(N^2)$ computational
cost, which is prohibitive for production-scale meshes. To resolve this
conflict, HGPTrans employs the slice attention mechanism of Transolver
\cite{ref_transolver}, which models the global dependencies in a fixed-size
latent space and thereby reduces the complexity to $O(N)$.

The slice attention proceeds in three steps. First, the vertex features are
projected onto a small set of $K$ learnable slices, where $K\ll N$.
Each vertex is softly assigned to the slices according to a learned score,
and a normalized weighted aggregation produces one token for each slice.
Intuitively, a slice groups vertices with similar learned states, and the
resulting token condenses their collective information. Second, a standard multi-head
self-attention is computed among the $K$ slice tokens, which is inexpensive
because $K$ is fixed and small. This attention stage explicitly models the
interaction between the slice tokens, thereby capturing the global aerodynamic
coupling between distant regions of the vehicle. Third, the updated slice
tokens are projected back onto the original vertices, distributing the global
context to each vertex while preserving its local identity. The detailed
formulation of the slice assignment, the normalized aggregation and the
projections is identical to that of the original Transolver
\cite{ref_transolver} and is therefore omitted here for brevity. In the
multi-head implementation, the slicing, normalized aggregation, attention, and
broadcasting operations are performed independently in each head before the
head outputs are concatenated. The aggregation of the slice tokens and their
projection back to the vertices each cost $O(KN)$,
whereas the attention among the $K$ slice tokens costs $O(K^2)$, so the
overall cost of the slice attention is $O(KN + K^2)$, which is linear in the
number of vertices because $K$ is a fixed constant. This design enables the
model to scale to the large meshes encountered in automotive CFD while
retaining a global receptive field.

The local GIN convolution and the global Transolver-style slice attention are combined in
an alternating manner within each HGPTrans block. The GIN layer provides the
local geometric context that is used to form the learned slices, and the
global attention writes the updated context back to the vertices, where a
subsequent local aggregation can refine the features. This alternating
local--global processing is repeated across the blocks, allowing the model to
build a progressively more comprehensive representation of the aerodynamic
behavior of the vehicle.

\subsection{Hierarchical graph pooling}\label{sec:pooling}

The third component of an HGPTrans block is a hierarchical graph-pooling layer
that progressively reduces the number of vertices. The pooling stage serves
two purposes: it decreases the computational cost of the subsequent layers, and
it induces a multi-scale representation of the vehicle surface. The vertex
selection criterion adopted here is based on the information content of each
vertex, following the node-information scoring principle introduced in
HGP-SL \cite{ref_hgpsl}.

The information score of a vertex measures how well its feature can be
reconstructed from the features of its neighbors. A vertex whose feature is
largely predictable from its neighborhood contains redundant information and
can be removed with little loss, whereas a vertex that cannot be reconstructed
from its neighbors is deemed informative and should be retained. Formally, the
information score $p_i$ of vertex $i$ is defined as the $L_1$ distance between
its feature and the mean-aggregated feature of its neighbors:
\begin{equation}
  p_i = \left\| \mathbf{h}_i - \left(\mathbf{D}^{-1}\mathbf{A}\mathbf{H}\right)_i
  \right\|_1,
\end{equation}
where $\mathbf{A}$, $\mathbf{D}$, and $\mathbf{H}$ are the adjacency matrix,
the degree matrix, and the node feature matrix of the graph entering the
pooling stage, respectively, and the $L_1$ norm is evaluated between the
feature vectors of vertex $i$ and of its mean-aggregated neighborhood. The
vertices of the graph are then
sorted by the corresponding information scores, and the top $rN$ vertices, where $r\in(0,1)$
is a predefined pooling ratio, are retained. Denoting by
$\mathrm{idx}$ the set of the retained vertex indices, the features and the
adjacency of the retained vertices form the coarsened graph
\begin{equation}
  \begin{aligned}
    \mathrm{idx} &= \mathrm{top\text{-}rank}\left(\mathbf{p},\, rN\right),\\
    \mathbf{H}' &= \mathbf{H}\left(\mathrm{idx}, :\right),\\
    \mathbf{A}' &= \mathbf{A}\left(\mathrm{idx}, \mathrm{idx}\right),
  \end{aligned}
\end{equation}
where $\mathrm{top\text{-}rank}$ returns the indices of the $rN$ vertices with
the largest scores, $\mathbf{H}(\mathrm{idx},:)$ collects the feature rows of
the retained vertices, and $\mathbf{A}(\mathrm{idx},\mathrm{idx})$ extracts
the submatrix of $\mathbf{A}$ indexed by the retained vertices, i.e., the
adjacency of the induced subgraph. In contrast to pooling methods based on a
single learnable
projection vector \cite{ref_graphunet,ref_sagpool} or on a soft cluster
assignment \cite{ref_diffpool}, the proposed criterion is parameter-free and
measures feature redundancy relative to the local neighborhood. It is intended
to assign relatively low scores to reconstructible vertices in smooth regions
and higher scores to less reconstructible vertices near geometric transitions.
The resulting selection therefore favors locally distinctive surface features
without assuming that every retained vertex is necessarily aerodynamic-critical.

The pooling layers are cascaded across the HGPTrans blocks so that the graph
is progressively coarsened from $N$ vertices to $rN$, then to $r^2N$, and so
on. In this work $L=4$ blocks are used, and the pooling ratio $r$ is set to
$0.8$. Both hyperparameters are determined by model selection on the
validation set, trading the prediction accuracy against the computational
cost, and their sensitivity is further analyzed in
Section~\ref{sec:param_abl}.
At each HGPTrans block, global mean and max pooling are applied to the retained
node features, and the two outputs are concatenated into a block-level graph
embedding. The block-level embeddings are then summed across the $L$ blocks and
fed into a fully connected regression head with a
$512$--$256$--$128$--$1$ structure that outputs the drag coefficient
$\hat{C}_D$.

\subsection{Comparison with prior methods}\label{sec:comparison}

Compared with voxel-based convolutional networks, which incur an $O(N^3)$
cost and lose surface detail \cite{ref_jacob,ref_voxnet}, with mesh-based
convolutional networks, which preserve the surface connectivity but operate
on the mesh elements \cite{ref_meshcnn}, and with point-cloud networks,
which disregard the mesh connectivity \cite{ref_pointnet,ref_drivaernet},
HGPTrans exploits the explicit surface topology through graph convolution. Compared with pure message-passing networks such as
MeshGraphNets \cite{ref_meshgraphnets} and Graph Network Simulators
\cite{ref_gns}, which require deep stacks to propagate information over long
distances, HGPTrans introduces a global slice attention that captures the
long-range aerodynamic coupling in a single operation. Compared with recent
graph-transformer surrogates such as AeroGTO \cite{ref_aerogto} and DoMINO
\cite{ref_domino}, HGPTrans additionally performs an explicit hierarchical
coarsening that is driven by the information content of the mesh vertices,
thereby reducing the computational cost and inducing a multi-scale
representation. This combination of local graph convolution, global
Transolver-style slice attention, and information-score-based hierarchical pooling
constitutes the main distinction of the proposed framework.

\section{Datasets and training strategy}\label{sec:dataset}

\subsection{Benchmark datasets: DrivAerNet and DrivAerNet++}\label{sec:drivaernet}

The proposed model is trained and evaluated on the two large-scale open-source
datasets DrivAerNet and DrivAerNet++. DrivAerNet \cite{ref_drivaernet} contains
approximately 4000 high-fidelity three-dimensional vehicle meshes generated by
parametric morphing of a reference geometry, each comprising on the order of
$4\times10^5$ surface faces. The dataset is multimodal, providing the
parametric description, the point cloud, the three-dimensional mesh, the
volume and surface flow fields, and the streamlines for every design, together
with the corresponding aerodynamic coefficients computed by high-fidelity CFD.
A distinctive feature of DrivAerNet is the explicit modeling of the wheels and
the underbody, which are often simplified or omitted in earlier datasets but
play a decisive role in the aerodynamic behavior of a real vehicle. DrivAerNet++
\cite{ref_drivaernetpp} extends this collection to approximately 8000 vehicle
designs, covering a wider range of body types, including fastback, notchback,
and estateback configurations, and a variety of underbody and wheel
geometries. This increased diversity provides a more challenging and realistic
benchmark for the evaluation of the generalization ability of the surrogate
model. Both datasets have been widely used to benchmark graph-based and
point-cloud-based drag prediction methods \cite{ref_drivaernet,
ref_drivaernetpp,ref_carbench}.

\subsection{Real-vehicle dataset}\label{sec:realcar}

\begin{figure}[tb]
  \centering
  \includegraphics[width=\linewidth]{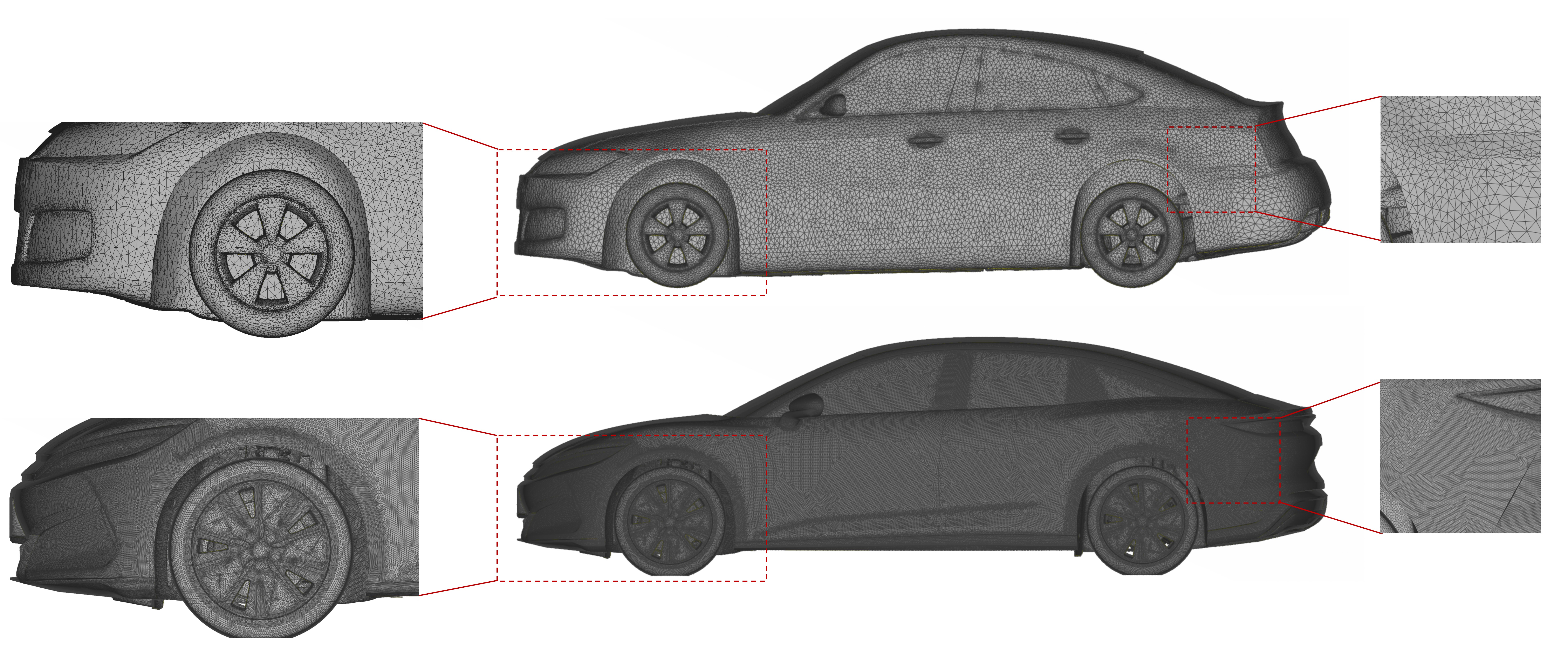}
  \caption{Comparison of representative vehicle geometries from the DrivAerNet
           open-source dataset and the real-vehicle dataset (a production
           sedan). The real-vehicle geometries contain a substantially higher
           level of geometric detail.}
  \label{fig:dataset}
\end{figure}

To assess the practical applicability of the proposed model, an in-house
real-vehicle dataset developed by BYD Auto is employed. The dataset contains
$400$ aerodynamic design schemes for mass-produced vehicles, including $200$
sedan samples and $200$ SUV samples, with five distinct vehicle models
covered in each category. The drag coefficient of the
samples spans a wide range from $0.19$ to $0.28$. This multi-model dataset
with a large $C_D$ span provides a heterogeneous setting for assessing the
cross-domain applicability of the model in an industrial scenario. Each design
is represented by a surface mesh containing on the order of $1\times10^7$
faces, which is about 25 times as many as the benchmark meshes, and the
real-vehicle geometries contain a large number of fine aerodynamic components,
such as front-wheel deflectors and air curtains, that are rarely present in
the parametric benchmark datasets. Fig.~\ref{fig:dataset} compares
representative samples of the DrivAerNet benchmark and the real-vehicle
dataset, illustrating the difference in the level of geometric detail.

\begin{figure}[tb]
  \centering
  \includegraphics[width=\linewidth]{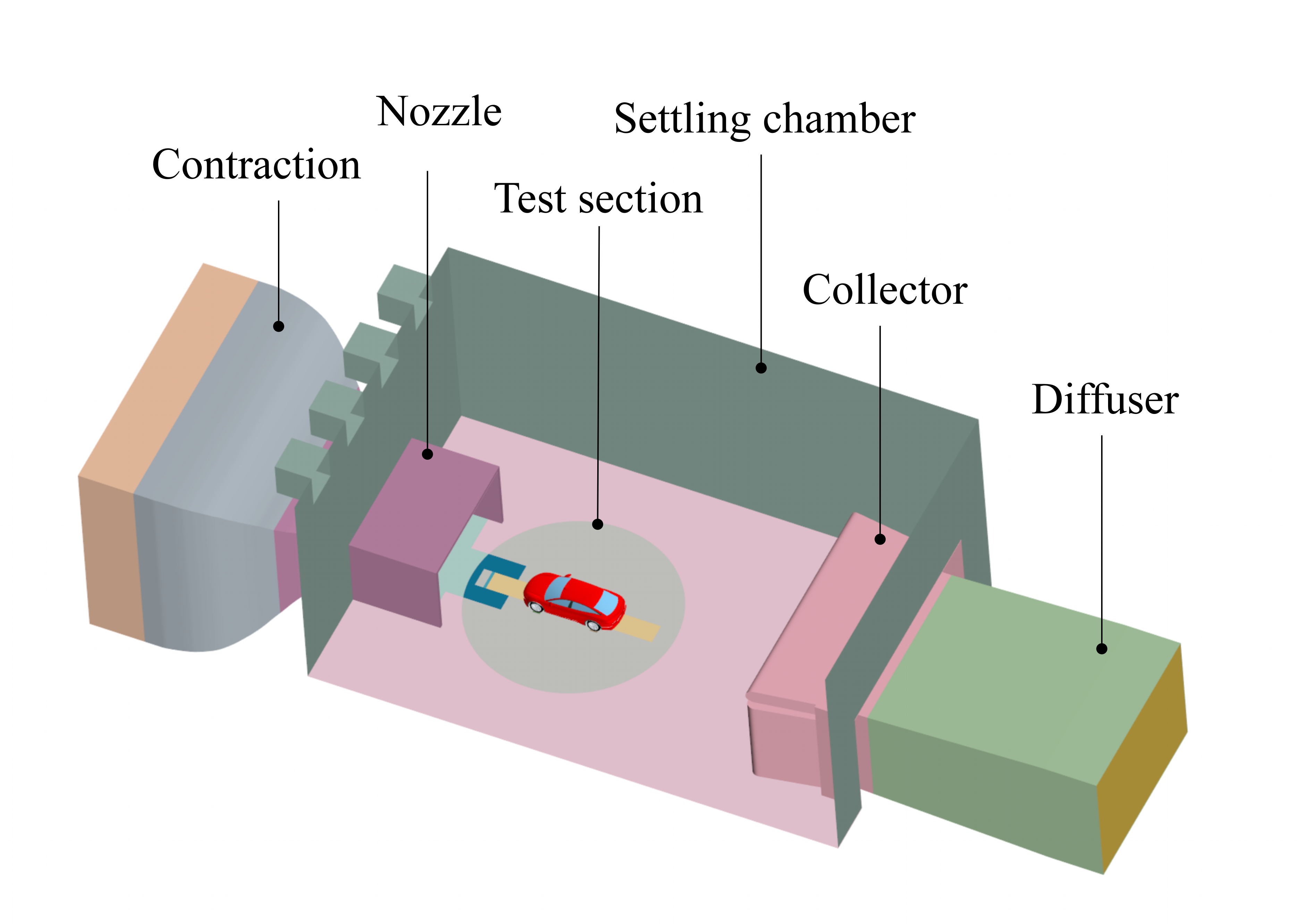}
  \caption{Computational domain of the full-scale digital wind
           tunnel.}
  \label{fig:wind_tunnel}
\end{figure}

The reference flow fields of the dataset are generated by steady
Reynolds-averaged Navier--Stokes (RANS) simulations in Simcenter STAR-CCM+ at
a freestream velocity of $U_\infty=120\,\mathrm{km\,h^{-1}}$. The
computational domain is a full-scale ($1{:}1$) digital reproduction of the
automotive wind tunnel at the China Automotive Engineering Research Institute
(CAERI) \cite{ref_caeri_drivaer}, as shown in Fig.~\ref{fig:wind_tunnel}. To limit the computational
cost, the closed-return circuit is truncated to the contraction, nozzle,
test section, collector and diffuser together with a downstream
extension; the nozzle exit measures $7\,\mathrm{m}\times4\,\mathrm{m}$ and the
test section is $18\,\mathrm{m}$ long. A velocity inlet
corresponding to $U_\infty$ is imposed at the upstream truncation plane of the
contraction and an outflow condition is applied at the downstream end of the
diffuser extension. The wind-tunnel walls and the stationary floor are treated
as stationary no-slip walls. The multistage boundary-layer control system
(BLCS) and the five-belt moving-ground system of the facility are reproduced
in the numerical model, so that ground motion and wheel rotation are simulated
simultaneously with the nozzle jet; the centre belt and the wheel-drive belts
are specified as moving no-slip walls and the wheels rotate according to a
rotating-wall condition. The domain is discretised with a
polyhedral/trimmed-cell mesh that combines a coarse background mesh with
multilevel local refinements in the open-jet shear layer, the near-ground
region and the vicinity of the vehicle; ten prism layers are generated on the
vehicle surface with $y^+\leq 2$ over the main body panels. The shear-stress
transport (SST) $k$--$\omega$ turbulence model is used with second-order
upwind discretisation and the all-$y^+$ wall treatment.

The drag force $F_D$ is obtained by integrating the streamwise pressure and
viscous forces over the vehicle surface, and the drag coefficient used as the
ground-truth label is
\begin{equation}
  C_D=\frac{F_D}{\tfrac{1}{2}\rho U_\infty^2 A_{\mathrm{ref}}},
\end{equation}
where $\rho$ is the air density and $A_{\mathrm{ref}}$ is the adopted frontal
reference area. Owing to the elevated mesh resolution and the presence of
complex fine structures, the real-vehicle dataset poses a substantially more
demanding task than the benchmark datasets, both in terms of prediction
accuracy and computational efficiency.

\subsection{Data preprocessing}\label{sec:preprocess}

For the benchmark datasets, the surface meshes are used directly. The surface
normals are computed from the mesh, and the six input channels (three
coordinates and three normal components) are standardized independently for
each mesh to zero mean and unit variance. For the real-vehicle dataset, the extremely dense meshes are first
simplified to a manageable resolution using the quadric error metric (QEM)
algorithm \cite{ref_qem}, which iteratively contracts the vertex pairs that
cause the smallest geometric error. In this work, the real-vehicle meshes are simplified to approximately $4\times10^5$ faces, the same order of magnitude as the number of faces of the DrivAerNet samples, before being fed into the network. This preprocessing reduces the computational cost and
the memory footprint while preserving the essential geometry of the body.

\subsection{Training and evaluation setup}\label{sec:setup}

Both benchmark and real-vehicle datasets are split into training, validation,
and test subsets with a fixed ratio of $70:15:15$. Because the GPU-memory
footprint of the large input graph permits only a single mesh per forward
pass, gradient accumulation over 16 successive samples is used, yielding an
effective batch size of 16. The model
minimizes the mean squared error (MSE) between the predicted and the
ground-truth drag coefficient using the Adam optimizer \cite{ref_adam}. The
initial learning rate is $5\times10^{-4}$, and the learning rate is halved
whenever the validation MSE has not improved for ten consecutive epochs,
following the \texttt{ReduceLROnPlateau} schedule implemented in
PyTorch \cite{ref_pytorch}. The network comprises
$L=4$ HGPTrans blocks
with a hidden dimension of $d=256$. The Transolver attention uses eight
attention heads and $K=32$ slices, and the pooling ratio is set to $r=0.8$. The
regression head is a fully connected network with three hidden layers of width
$512$, $256$, and $128$, followed by a scalar output layer. Both the training
and the inference are carried
out on NVIDIA L20 GPUs, and the inference time reported throughout this paper
is measured on a single L20 GPU. The reported inference time is the average
per-vehicle end-to-end processing time, including data loading and mesh
simplification, computed over the entire test set with a batch size of one.

The prediction performance is measured by the mean absolute error (MAE), the
mean squared error (MSE), the maximum absolute error, and the coefficient of
determination $R^2$. For $n$ samples with ground-truth and predicted drag
coefficients $C_{D,i}$ and $\hat{C}_{D,i}$, respectively, the reported relative
$L_1$ and $L_2$ errors are defined as
\begin{equation}
  \begin{aligned}
    E_{L_1}^{\mathrm{rel}}
      &=\frac{\sum_{i=1}^{n}|\hat{C}_{D,i}-C_{D,i}|}
              {\sum_{i=1}^{n}|C_{D,i}|}\times100\%,\\
    E_{L_2}^{\mathrm{rel}}
      &=\sqrt{\frac{\sum_{i=1}^{n}(\hat{C}_{D,i}-C_{D,i})^2}
                    {\sum_{i=1}^{n}C_{D,i}^2}}\times100\%.
  \end{aligned}
\end{equation}
The parameter count and the inference time are also reported. For the
real-vehicle experiments, transfer learning is adopted: the model is first
pre-trained on DrivAerNet++ and then fine-tuned on a small subset of the
real-vehicle data. This strategy
leverages the general aerodynamic features learned from the synthetic
parametric geometries and substantially reduces the amount of real-vehicle
data required to reach a satisfactory accuracy.

\section{Experimental results and analysis}\label{sec:experiments}

\subsection{Prediction performance on DrivAerNet and DrivAerNet++}\label{sec:benchmark}

The prediction performance of HGPTrans on the DrivAerNet and DrivAerNet++
datasets is compared with available results from mainstream baseline models,
including
point-cloud networks (PointNet \cite{ref_pointnet}, PointNet++
\cite{ref_pointnetpp}, PointNeXt \cite{ref_pointnext}, ASSANet
\cite{ref_assanet}, and Point-BERT \cite{ref_pointbert}), graph networks
(GCNN \cite{ref_drivaernet}, RegDGCNN \cite{ref_drivaernet}, DeepGCNs \cite{ref_deepgcn}, and
MeshGraphNets \cite{ref_meshgraphnets}), convolution network (FIGConvNet
\cite{ref_figconv}), and triplane network TripNet \cite{ref_tripnet}. Table~\ref{tab:main}
reports eleven baseline results for DrivAerNet and four available baseline
results for DrivAerNet++. It lists the mean squared error, the mean absolute
error, the maximum absolute error, and the coefficient of determination.

\begin{table*}[tb]
\caption{Comparison of HGPTrans with mainstream baseline models on the
           DrivAerNet and DrivAerNet++ datasets. The best value within each
           dataset block is highlighted in bold. Baseline entries are taken
           from the cited literature and use the splits reported by the
           respective sources.}
  \label{tab:main}
  \centering
  \small
  \begin{tabular}{@{}llccccc@{}}
    \toprule
    Dataset & Model & MSE ($10^{-5}$) $\downarrow$ & MAE ($10^{-3}$) $\downarrow$
             & MaxAE ($10^{-2}$) $\downarrow$ & $R^2$ $\uparrow$ \\
    \midrule
    & \textbf{HGPTrans} & \textbf{2.398} & \textbf{3.718} & 1.936 & 0.968 \\
    & TripNet & 2.602 & 4.030 & \textbf{1.268} & \textbf{0.972} \\
    & FIGConvNet & 3.225 & 4.423 & 2.134 & 0.957 \\
    & PointNeXt & 4.577 & 5.200 & 2.410 & 0.939 \\
    & ASSANet & 5.433 & 5.810 & 2.390 & 0.927 \\
    & MeshGraphNets & 6.000 & 6.080 & 2.965 & 0.917 \\
    DrivAerNet & DeepGCNs & 6.297 & 6.091 & 3.070 & 0.916 \\
    & Point-BERT & 6.334 & 6.204 & 2.767 & 0.915 \\
    & PointNet++ & 7.813 & 6.755 & 3.463 & 0.896 \\
    & RegDGCNN & 8.010 & 6.910 & 8.800 & 0.901 \\
    & GCNN & 10.700 & 7.170 & 10.970 & 0.874 \\
    & PointNet & 12.000 & 8.850 & 10.180 & 0.826 \\
    \midrule
    & \textbf{HGPTrans} & \textbf{5.625} & \textbf{5.694} & \textbf{3.054} & 0.864 \\
    & TripNet & 9.100 & 7.170 & 7.700 & \textbf{0.957} \\
    & RegDGCNN & 14.200 & 9.310 & 12.790 & 0.641 \\
    DrivAerNet++ & PointNet & 14.900 & 9.600 & 12.450 & 0.643 \\
    & GCNN & 17.100 & 10.430 & 15.030 & 0.596 \\
    \bottomrule
  \end{tabular}
\end{table*}

\begin{figure*}[tb]
  \centering
  \includegraphics[width=\textwidth]{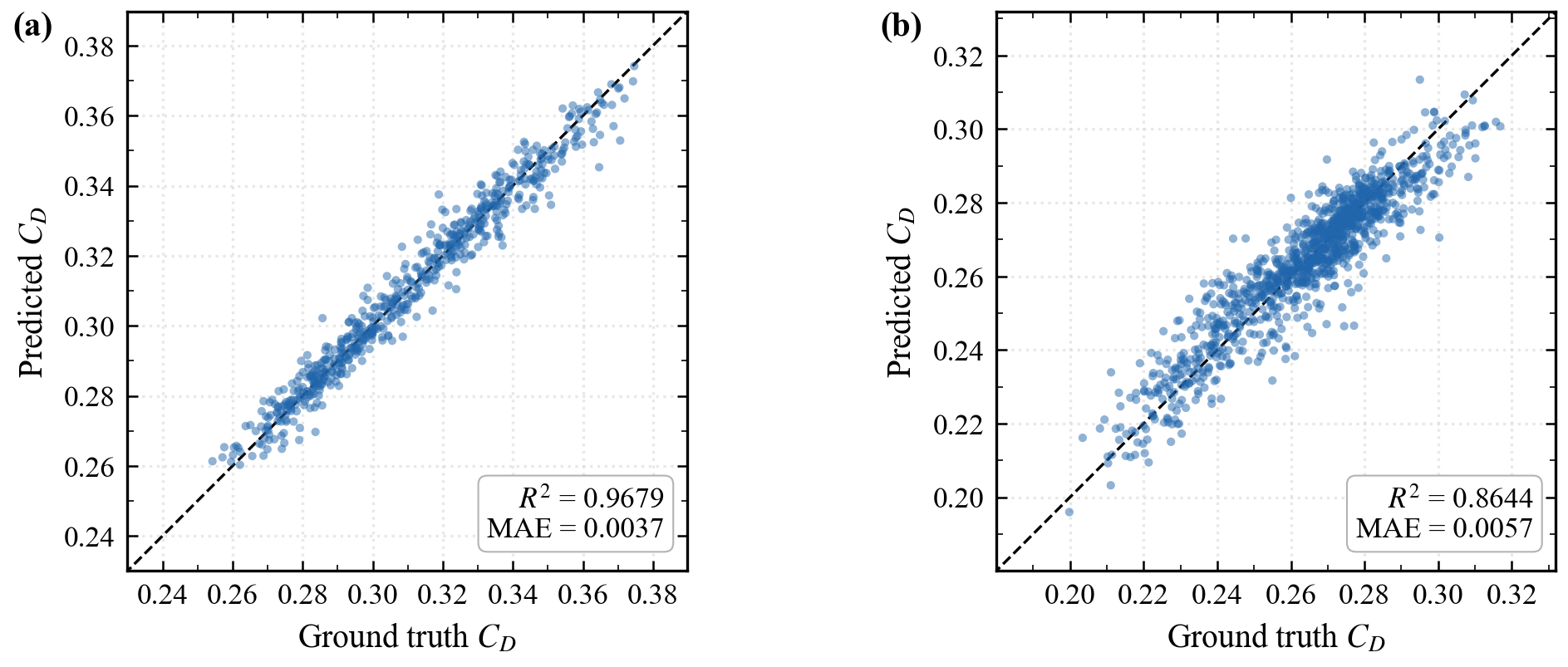}
  \caption{Scatter plots of the predicted versus the ground-truth drag
           coefficient on the test sets of (a) DrivAerNet and
           (b) DrivAerNet++. The dashed line indicates the ideal $y=x$
           diagonal.}
  \label{fig:scatter}
\end{figure*}

On DrivAerNet, HGPTrans reports an MSE of $2.398\times10^{-5}$ and an MAE of
$3.718\times10^{-3}$, both numerically lower than the available baseline values
listed in Table~\ref{tab:main}, and its
coefficient of determination $R^2=0.968$ is only slightly below that of
TripNet ($0.972$). The advantage of HGPTrans over the graph-based baselines is
substantial: compared with RegDGCNN, the MAE is reduced by approximately
$46\%$, and compared with GCNN and PointNet, the reduction exceeds $48\%$.
These results indicate that the hierarchical local--global architecture is
effective in capturing the aerodynamic behavior of the vehicle from the
surface mesh. Fig.~\ref{fig:scatter} shows the scatter plot of the predicted
versus the ground-truth $C_D$ on the two benchmark datasets, in which most
samples lie close to the diagonal line, indicating a high prediction accuracy
across the drag-coefficient distribution.

On the more challenging DrivAerNet++ dataset, HGPTrans reports an MSE of
$5.625\times10^{-5}$, an MAE of $5.694\times10^{-3}$, and a maximum absolute
error of $3.054\times10^{-2}$; all three values are numerically lower than the
available values listed in Table~\ref{tab:main}. Relative to the reported
TripNet values, the MAE is lower
by approximately $21\%$ and the MSE by approximately $38\%$. The relative $L_1$ and $L_2$
errors of the model on DrivAerNet++ are $2.15\%$ and $2.83\%$, respectively,
indicating that the majority of the predictions are within a few percentage
points of the ground truth. As shown in Fig.~\ref{fig:scatter}(b), the ground-truth drag coefficients of
the test samples span a wide range, from $0.20$ to $0.32$, and although the
points remain
clustered around the ideal diagonal, the scatter is visibly broader than in
Fig.~\ref{fig:scatter}(a). The higher absolute error on DrivAerNet++ compared
with DrivAerNet is expected, because the former dataset spans a wider range of
body types and geometric configurations, making the prediction task inherently
more difficult.

In addition to the prediction accuracy, the computational efficiency of the
model is evaluated. HGPTrans comprises approximately $2.490$ million trainable
parameters. During inference, the model predicts the drag coefficient of a
single DrivAerNet-scale vehicle in about $0.21$ s per sample on an NVIDIA L20
GPU, averaged over the entire test set. This inference time is several orders
of magnitude shorter than the hours required by a high-fidelity CFD
simulation, demonstrating the suitability of the model for the rapid screening
of a large number of design candidates during the early styling phase.
An analysis of the prediction error distribution further reveals that on
DrivAerNet, $81.2\%$ of the test samples are predicted with a relative error
below $2\%$ with respect to the CFD reference and $94.5\%$ below $3\%$; on
DrivAerNet++, the corresponding proportions are $56.1\%$ and $75.3\%$,
respectively, confirming that the model delivers consistently tight
predictions for the majority of the test distribution.

\subsection{Transfer learning on the real-vehicle dataset}\label{sec:transfer}

\begin{figure*}[tb]
  \centering
  \includegraphics[width=\textwidth]{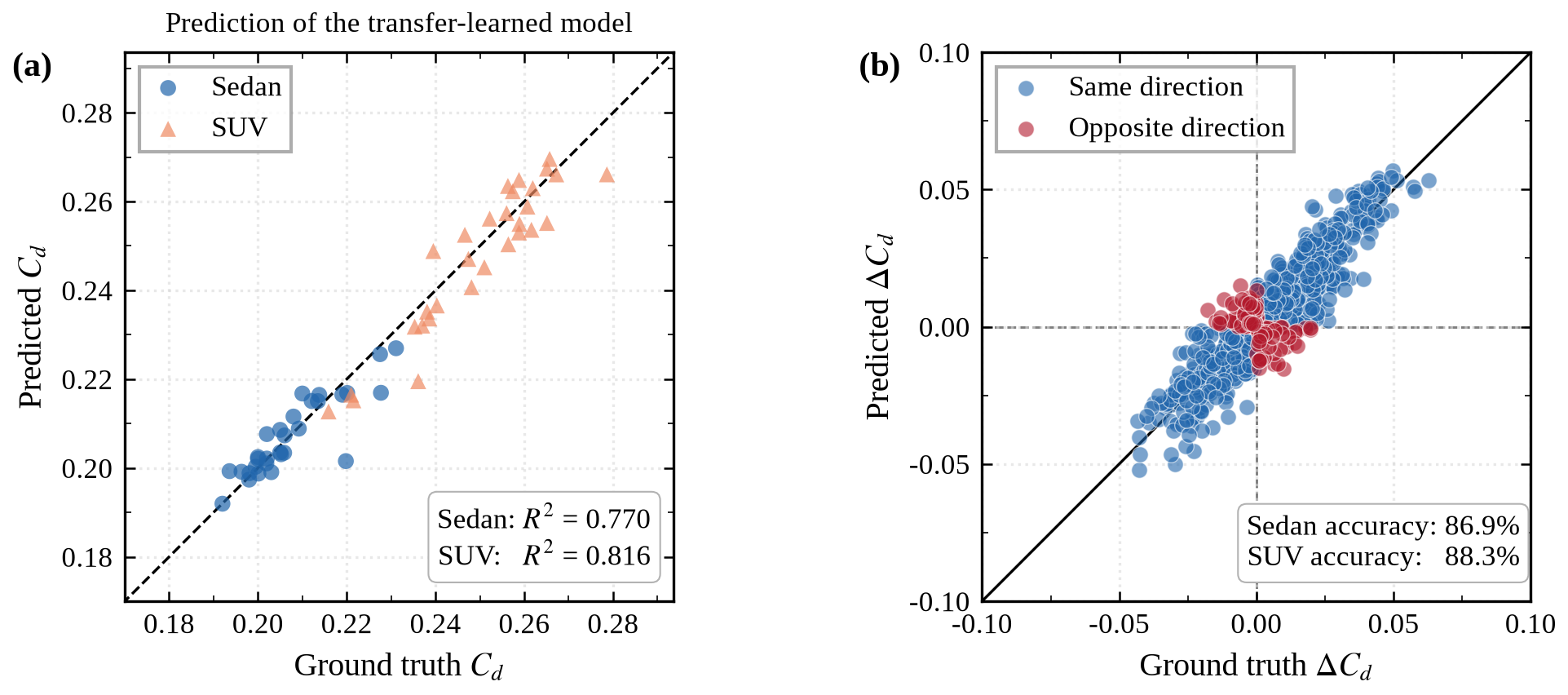}
  \caption{(a) Scatter plot of the predicted versus the ground-truth drag
           coefficient on the real-vehicle test set; the inset reports the
           coefficient of determination for the sedan and SUV subsets.
           (b) Scatter plot of the predicted versus the ground-truth
           difference $\Delta C_D$ for all pairs of variants within the sedan
           and SUV subsets; blue and red markers denote the pairs for which
           the model predicts the direction of the change correctly and
           incorrectly, respectively, and the inset reports the direction
           accuracy for each subset.}
  \label{fig:realcar}
\end{figure*}

The cross-domain applicability of the model to the real-vehicle dataset is evaluated
through transfer learning. The model is first pre-trained on DrivAerNet++ and
then fine-tuned on the real-vehicle data, with a single unified model covering
both the sedan and SUV categories, as described in
Section~\ref{sec:setup}. Owing to the difference in mesh resolution and the
presence of fine aerodynamic components, the real-vehicle meshes are first
simplified with the QEM algorithm, and the model is then fine-tuned on the
simplified meshes. Fig.~\ref{fig:realcar} shows the scatter plots of the
predicted versus the ground-truth drag coefficient and of the predicted versus
the ground-truth drag-coefficient difference for all unordered pairs of test
variants within the sedan and SUV subsets. Let $\mathcal{P}$ denote this set of
within-category pairs, and define $\Delta C_{D,ij}=C_{D,i}-C_{D,j}$ and
$\Delta\hat{C}_{D,ij}=\hat{C}_{D,i}-\hat{C}_{D,j}$. The pairwise direction
accuracy is defined as
\begin{equation}
  A_{\mathrm{dir}}=
  \frac{1}{|\mathcal{P}|}\sum_{(i,j)\in\mathcal{P}}
  \mathbb{I}\!\left[\Delta\hat{C}_{D,ij}\Delta C_{D,ij}>0\right]
  \times100\%,
\end{equation}
where $\mathbb{I}[\cdot]$ is the indicator function. The pairs are constructed
separately within the sedan and SUV subsets, and the direction accuracy is
reported for each subset.

After transfer learning from the DrivAerNet++ pre-training, the model
attains a relative $L_1$ error of $1.56\%$ for sedans and $2.12\%$ for SUVs,
with MAE of $3.227\times10^{-3}$ and $5.306\times10^{-3}$ and
$R^2$ of $0.770$ and $0.816$, respectively. The
residual errors are larger for the SUV subset, which covers a higher and more
widely spread range of $C_D$ values. The drag coefficient is predicted with a
relative error below $3\%$ for $90.0\%$ of the sedan variants and $83.3\%$ of
the SUV variants. As shown in Fig.~\ref{fig:realcar}(b), the direction of the
drag change is predicted correctly for $86.9\%$ of the sedan variant pairs and
$88.3\%$ of the SUV variant pairs. The inference time for a single
real-vehicle design is approximately $0.293$ s on a single L20 GPU.
Table~\ref{tab:transfer} summarizes the transfer-learning results for the two
subsets.

\begin{table*}[htb]
  \caption{The results of HGPTrans on the real-vehicle dataset,
           evaluated on the held-out test set and reported for the total test
           set and the sedan and SUV subsets. The upper block corresponds to
           the proposed transfer-learning setting (pre-trained on
           DrivAerNet++); the lower block is a baseline trained from scratch
           on the real-vehicle data only.}
  \label{tab:transfer}
  \centering
  \footnotesize
  \begin{tabular}{@{}lcccccc@{}}
    \toprule
    Subset & MSE ($10^{-5}$) $\downarrow$ & MAE ($10^{-3}$) $\downarrow$
      & MaxAE ($10^{-2}$) $\downarrow$ & Rel. $L_1$ ($\%$) $\downarrow$
      & Rel. $L_2$ ($\%$) $\downarrow$ & $R^2$ $\uparrow$ \\
    \midrule
    \multicolumn{7}{@{}l}{\emph{With transfer learning (pre-trained on DrivAerNet++)}}\\
    Total & 3.152 & 4.267 & 1.826 & 1.865 & 2.441 & / \\
    Sedan & 2.311 & 3.227 & 1.826 & 1.555 & 2.314 & 0.770 \\
    SUV   & 3.993 & 5.306 & 1.656 & 2.123 & 2.524 & 0.816 \\
    \midrule
    \multicolumn{7}{@{}l}{\emph{Without transfer learning (trained from scratch on the real-vehicle data)}}\\
    Total & 5.549 & 5.341 & 3.405 & 2.335 & 3.238 & / \\
    Sedan & 4.048 & 4.841 & 1.643 & 2.333 & 3.063 & 0.598 \\
    SUV   & 7.056 & 5.850 & 3.405 & 2.341 & 3.355 & 0.675 \\
    \bottomrule
  \end{tabular}
\end{table*}

The comparison with the
baseline trained from scratch on the real-vehicle data alone quantifies the
benefit of the transfer-learning strategy: pre-training on DrivAerNet++
reduces the total MSE from $5.549\times10^{-5}$ to $3.152\times10^{-5}$ and,
in particular, raises the per-class $R^2$ from $0.598$ and $0.675$ to $0.770$
and $0.816$ for the sedan and SUV subsets, respectively. The from-scratch
baseline also reduces the per-class accuracy gap only marginally, which
indicates that the small real-vehicle training set is insufficient to learn
the geometry--aerodynamics relationship without a large-scale prior.
Compared with the high-fidelity CFD
simulation, which requires on the order of tens of hours for such dense
geometries, the model achieves a speedup of approximately five orders of
magnitude. This result demonstrates that, after a small amount of fine-tuning,
the model can provide an accurate and rapid estimate of the drag coefficient
for production vehicle geometries, and that it can preserve the direction of
the drag difference for most pairs drawn from the same broad vehicle category.

The observed transfer performance may be associated with two aspects. First,
pre-training on the large-scale DrivAerNet++ dataset may provide the network
with a useful prior about the
relationship between the vehicle geometry and the aerodynamic behavior. This
prior is transferred to the real-vehicle domain by the fine-tuning stage, which
adapts the pretrained representation to the real-vehicle CFD data. Second, the
QEM-based simplification of the dense real-vehicle meshes is intended to align
the input resolution with that seen during pre-training and may reduce the
distribution shift between the source and the target domain.

Despite the encouraging accuracy, the absolute drag values predicted by the
model should be interpreted with care. As reported in the real-world benchmark
study \cite{ref_realworld}, deep surrogates tend to capture the broad trends
across different vehicle families more reliably than the small intra-family
variations, and the model is most valuable as a fast screening tool that ranks
candidate designs rather than as a replacement for the final validation CFD.
In the present experiments, the within-category pairwise result supports the
use of the model for early screening among candidates in the sedan and SUV
groups. Nevertheless, additional high-fidelity evaluations remain necessary
before a design is frozen.

\subsection{Ablation studies}\label{sec:ablation}

\subsubsection{Architecture ablation}\label{sec:arch_abl}

To quantify the contribution of each architectural component, a series of
ablation experiments is conducted by replacing or removing the individual
modules of HGPTrans. The results are reported in Table~\ref{tab:abl}. The full
model serves as the baseline, and each variant modifies a single component
while keeping the remaining architecture unchanged.

\begin{table*}[tb]
  \caption{Architecture ablation study on DrivAerNet. The full model is
           compared with variants in which the local convolution, the global
           attention, the pooling, and the input features are modified.}
  \label{tab:abl}
  \centering
  \small
  \begin{tabular}{@{}lcccc@{}}
    \toprule
    Variant & MSE ($10^{-5}$) & MAE ($10^{-3}$) & $R^2$ & Param. ($10^6$)  \\
    \midrule
    \textbf{Full HGPTrans} & \textbf{2.398} & \textbf{3.718} & \textbf{0.968}
             & 2.490\\
    GIN $\rightarrow$ GCN   & 2.445 & 3.781 & 0.967 & 2.226 \\
    GIN $\rightarrow$ MLP   & 2.635 & 3.938 & 0.965 & 2.490  \\
    w/o Transolver          & 2.481 & 3.820 & 0.967 & 0.628 \\
    w/o pooling             & 2.726 & 4.021 & 0.964 & 2.488 \\
    w/o normal features     & 2.866 & 4.153 & 0.962 & 2.489 \\
    \bottomrule
  \end{tabular}
\end{table*}

\begin{itemize}
  \item \textbf{Role of the GIN convolution.} Replacing the GIN layer with a
        standard GCN, which uses degree-normalized aggregation instead of the
        unnormalized sum aggregation of GIN, increases the MAE from
        $3.718\times10^{-3}$ to
        $3.781\times10^{-3}$. Replacing it with a plain MLP that does not
        consider the graph connectivity at all degrades the MAE further to
        $3.938\times10^{-3}$. These results confirm that the graph-aware sum
        aggregation appears to provide a more discriminative local geometric encoding
        than both the degree-normalized GCN aggregation and the
        connectivity-agnostic MLP.
  \item \textbf{Role of the Transolver attention.} Removing the Transolver
        attention module reduces the parameter count from $2.490$ million to
        $0.628$ million, yet the MSE increases by $3.5\%$ (from
        $2.398\times10^{-5}$ to $2.481\times10^{-5}$). The accuracy loss
        incurred by removing a module that accounts for most of the parameters
        indicates that the global slice attention provides information that is
        not fully recovered by the local GIN convolution alone.
  \item \textbf{Role of the hierarchical pooling.} Removing the pooling layer
        increases the MAE from $3.718\times10^{-3}$ to $4.021\times10^{-3}$. This result indicates that
        the information-score-based coarsening is beneficial to the observed
        prediction accuracy and reduces the number of nodes processed by the
        subsequent blocks.
  \item \textbf{Role of the input features.} When the surface-normal
        components are removed from the node features, the MAE increases to
        $4.153\times10^{-3}$, the largest degradation among all the ablations.
        This indicates that the local orientation information encoded by the
        normals is essential for describing the geometry of the vehicle
        surface, and that the model relies heavily on this feature for the drag
        prediction.
\end{itemize}

\subsubsection{Hyperparameter ablation}\label{sec:param_abl}

\begin{figure}[tb]
  \centering
  \includegraphics[width=\linewidth]{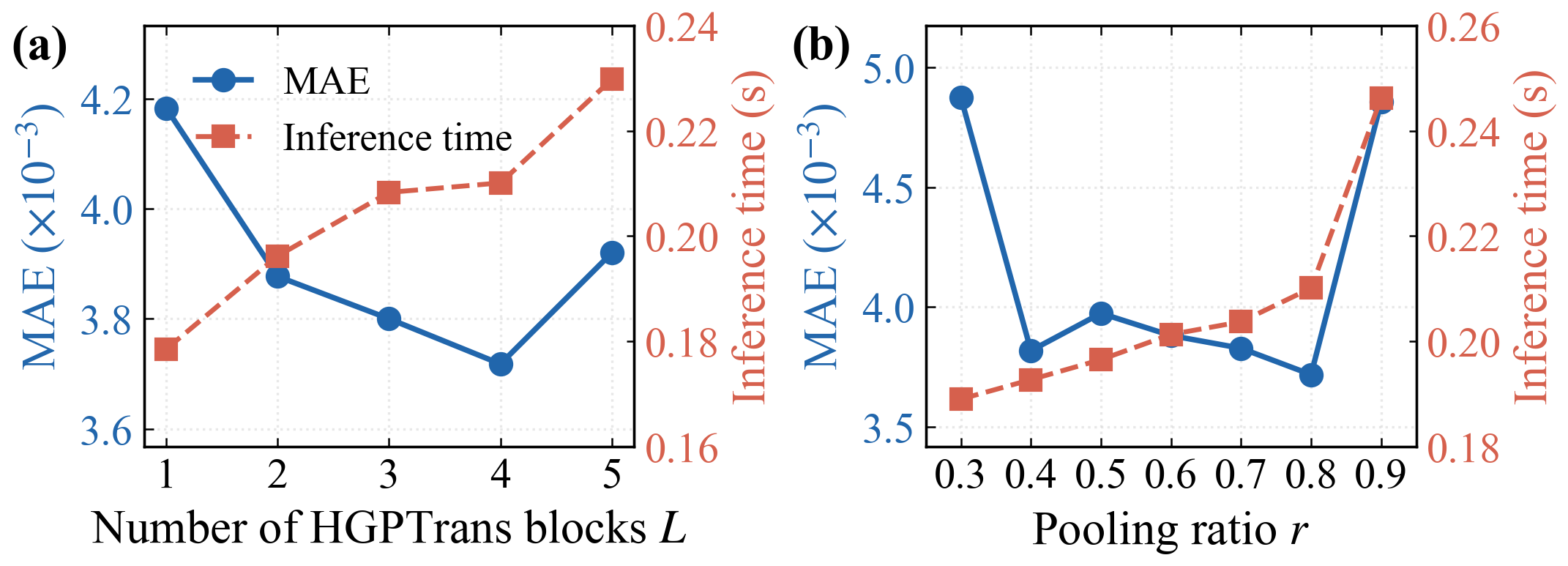}
  \caption{Hyperparameter ablation on DrivAerNet: (a) effect of the number of
           HGPTrans blocks $L$ and (b) effect of the pooling ratio $r$ on the
           mean absolute error and the inference time.}
  \label{fig:abl}
\end{figure}

The sensitivity of the model to the network depth and the pooling ratio is
investigated. As stated in Sections~\ref{sec:pooling} and \ref{sec:setup},
$L=4$ and $r=0.8$ are determined by model selection on the validation set;
the following analysis examines how the performance responds to these two
hyperparameters and confirms that the selected values lie near the optimum.
Fig.~\ref{fig:abl}(a) shows the prediction performance and the
inference time as a function of the number of HGPTrans blocks $L$, and
Fig.~\ref{fig:abl}(b) shows the corresponding results as a function of the
pooling ratio $r$.

As shown in Fig.~\ref{fig:abl}(a), the MAE decreases monotonically from
$4.183\times10^{-3}$ at $L=1$ to its lowest value of $3.718\times10^{-3}$ at
$L=4$, and increases to $3.921\times10^{-3}$ when a fifth block is added. The inference
time increases monotonically with the number of blocks, from $0.179$~s at
$L=1$ to $0.230$~s at $L=5$. Accordingly, $L=4$ is selected as a balance
between the prediction accuracy and the computational cost. As shown in
Fig.~\ref{fig:abl}(b), the pooling ratio $r$ controls the trade-off between the
retained information and the degree of compression. A small pooling ratio
($r=0.3$) removes too many vertices and leads to a substantial increase of the
MAE to $4.875\times10^{-3}$, because the aerodynamic-critical features are
discarded. A large pooling ratio ($r=0.9$), on the other hand, retains almost
all of the vertices, so that the inference cost is the highest of the sweep
($0.246$~s per sample) while the MAE nevertheless increases to
$4.856\times10^{-3}$. The best performance is obtained at $r=0.8$
($3.718\times10^{-3}$ at $0.210$~s per sample), which is therefore adopted in
the final model.

Taken together, the ablation studies support the design rationale of HGPTrans:
the sum-aggregation graph convolution, the global slice attention, the
information-score-based hierarchical pooling, and the inclusion of the surface
normals all contribute to the accurate prediction of the drag coefficient,
while the hierarchical coarsening additionally reduces the computational cost.
Moreover, the accuracy gains of the global attention and of the pooling are
not merely a consequence of an increased parameter count: removing the
Transolver attention reduces the parameter count to $0.63$ million yet
degrades the accuracy, whereas removing the pooling changes the parameter count
only slightly but still increases the error. These observations suggest that the
architectural inductive biases of HGPTrans contribute to the observed
performance beyond the effect of parameter count alone.

\section{Conclusions}\label{sec:conclusions}

This paper proposed HGPTrans, a hierarchical graph-pooling network combined
with a Transolver-style slice attention mechanism, for the direct prediction
of the automotive drag coefficient from the vehicle surface mesh. By
alternating a graph isomorphism convolution for local geometry, a slice
attention for global aerodynamic coupling, and an information-score-based
hierarchical pooling for progressive graph coarsening, the model achieves the
lowest mean absolute and mean squared errors among the baseline results listed
for DrivAerNet and DrivAerNet++ in Table~\ref{tab:main}, while requiring
approximately $2.490$ million parameters. Through transfer learning, the representation
learned from synthetic parametric geometries can be adapted to production
vehicles covering both sedans and SUVs,
attaining relative $L_1$ errors of $1.56\%$ for sedans and $2.12\%$ for SUVs
on the real-vehicle test set, correctly predicting the direction of the drag
change for $86.9\%$ of the sedan variant pairs and $88.3\%$ of the SUV variant
pairs, at about $0.293$ s per
vehicle---several orders of magnitude faster than high-fidelity CFD.
The ablation studies further indicate that the sum-aggregation convolution, the
global slice attention, the hierarchical pooling, and the surface-normal
features each contribute to the accuracy.

Despite these promising results, the current work has several limitations. The
model is evaluated on steady-state external flow and rigid geometries, and its
behavior under transient conditions, such as yaw and crosswind, is not
considered. In addition, the real-vehicle evaluation is confined to the
production models of a single manufacturer, and the generalization to active
aerodynamic components, such as movable grilles and spoilers, remains to be
investigated. Future work will extend the model to transient conditions,
active flow-control components, and the coupled prediction of the surface
pressure and shear-stress fields, as well as to the integration of the
surrogate into a closed-loop aerodynamic optimization framework.

\begin{acknowledgments}
This work was supported by the Shenzhen Science and Technology Program
(Grant No. ZDCYKCX20250901092759009).
\end{acknowledgments}

\section*{Declarations}
\noindent\textbf{Conflict of interest} On behalf of all the authors, the
corresponding author states that there is no conflict of interest.

\medskip
\noindent\textbf{Author contributions}
\textbf{Bo Liu}: conceptualization, methodology, software,
writing -- original draft.\quad \textbf{Fengli Zhang}: resources, writing -- review \& editing.\quad \textbf{Qiuli Luo}: data curation,
validation, visualization.\quad \textbf{Lianrui Nie}: investigation, formal analysis.\quad
\textbf{Wenjiang Wang}: funding
acquisition, writing -- review \& editing.

\medskip
\noindent\textbf{Data availability} The DrivAerNet and DrivAerNet++ datasets
used in this study are openly available to the research community. The
real-vehicle dataset is proprietary to BYD Auto Industry Co., Ltd.\ and
contains confidential vehicle design information; it is therefore not
publicly available and cannot be shared. The source code is publicly
available at \url{https://github.com/BoLiu-USTC/HGPTrans}.

\renewcommand{\refname}{References} 

\bibliography{refs}

\end{document}